\documentclass[sigconf,natbib=false, nonacm]{acmart}

\AtBeginDocument{%
  \providecommand\BibTeX{{%
    Bib\TeX}}}

\graphicspath{{images/}}

\def\BibTeX{{\rm B\kern-.05em{\sc i\kern-.025em b}\kern-.08em
    T\kern-.1667em\lower.7ex\hbox{E}\kern-.125emX}}

\usepackage{amsmath,amsfonts}
\usepackage{algorithmic}
\usepackage{array}
\usepackage{booktabs}       
\usepackage{comment}
\usepackage{nicefrac}       
\usepackage{microtype}      
\usepackage{textcomp}
\usepackage{stfloats}
\usepackage{url}
\usepackage{verbatim}
\usepackage{graphicx}
\usepackage[export]{adjustbox}
\usepackage{subcaption}
\usepackage{enumitem}
\usepackage{tikz}
\usepackage{xcolor}
\usepackage{balance}
\usepackage[shortcuts]{extdash}
\newcommand{\mycomment}[1]{}
\newcommand*\circled[1]{\tikz[baseline=(char.base)]{
            \node[shape=circle,draw,inner sep=.75pt] (char) {#1};}}

\RequirePackage[
  datamodel=acmdatamodel,
  style=acmnumeric,
  ]{biblatex}

\begin{document}

\title{SAP: Syntactic Attention Pruning for Transformer-based Language Models}

\author{Tzu-Yun Lee}
\email{grace6029@iis.sinica.edu.tw}
\affiliation{%
  \institution{Institute of Information Science, Academia Sinica}
  \city{Taipei}
  \country{Taiwan}
}

\author{Ding-Yong	Hong}
\email{dyhong@iis.sinica.edu.tw}
\affiliation{%
  \institution{Institute of Information Science, Academia Sinica}
  \city{Taipei}
  \country{Taiwan}
}

\author{Jan-Jan	Wu}
\email{wuj@iis.sinica.edu.tw}
\affiliation{%
  \institution{Institute of Information Science, Academia Sinica}
  \city{Taipei}
  \country{Taiwan}
}

\renewcommand{\shortauthors}{Tzu-Yun Lee et al.}

\begin{abstract}
This paper introduces Syntactic Attention Pruning (SAP), a novel method for effectively pruning attention heads in Transformer models. 
Unlike conventional approaches that rely solely on mathematical analysis of model weights and activations, SAP incorporates both the syntactic structure and attention patterns of sentences to guide the pruning process.
By leveraging these linguistic features, SAP not only achieves performance comparable to state-of-the-art methods but also enhances the interpretability of model behavior.
To further improve robustness, we propose Candidate Filtering (CF), a mechanism that prioritizes heads based on their contribution to model performance, mitigating degradation during pruning.
Experimental results indicate that SAP effectively preserves critical heads of a high density of strong attention values, outperforming existing head pruning strategies in retrain-free settings. 
These findings position SAP as a promising foundation for a new direction in model compression research, offering high flexibility for pruning across all transformer-based language models.
\end{abstract}

\ccsdesc[500]{Computing methodologies~Natural language processing}
\ccsdesc[500]{Computing methodologies~Neural networks}

\keywords{Model Compression, Natural Language Processing, Deep Learning}

\maketitle

\section{Introduction}  \label{introduction}
Since the introduction of the Transformer architecture~\cite{NIPS2017_3f5ee243}, transformer\-/based language models have rapidly scaled in both parameter count and supported context length, leading to significant advances in artificial intelligence across a wide range of tasks.
However, these advances come at the cost of growing computational demands, with inference increasingly constrained by memory bandwidth due to the overhead of parameter storage~\cite{yuan2024llminferenceunveiledsurvey}.
For example, inference with LLaMA-7B~\cite{touvron2023llama} on an input sequence of 1,000 tokens requires approximately 7,000 trillion floating-point operations and 28 GB of memory to store the model parameters in FP32 precision.
Consequently, mitigating these computational and memory bottlenecks is essential for achieving scalable and efficient deployment.

Model compression is a widely adopted technique to reduce the size of deep learning models.
Common compression techniques include quantization~\cite{pmlr-v139-kim21d, NEURIPS2022_adf7fa39}, knowledge distillation~\cite{distillation}, and weight pruning~\cite{10.1016/j.eswa.2025.126957, 10.5555/2969239.2969366, sun2023wanda}.
Weight pruning removes redundant or unimportant weights from a neural network, offering
several key benefits such as improved inference efficiency, reduced memory footprint, and, in some cases, enhanced model generalization and robustness.

The core challenge in weight pruning lies in accurately identifying which weights contribute least to model performance, aiming to minimize accuracy degradation after pruning.
Most pruning techniques rely on {\em mathematical metrics} to estimate weight importance. Researchers have proposed a variety of methods to guide the pruning process, including magnitude-based~\cite{10.1016/j.eswa.2025.126957, 10.5555/2969239.2969366, sun2023wanda}, gradient-based~\cite{NEURIPS2019_2c601ad9, das2024sizegradientsshapepruning, alizadeh2022prospectpruningfindingtrainable}, and activation-based criteria~\cite{wang-etal-2025-cfsp, shen-etal-2024-pruning}.
Magnitude-based methods remove weights with the smallest magnitude, while activation-based methods prune weights connected to neurons with low activation values, based on the premise that lower values of weights or activations contribute less to the model's predictive capacity.
In contrast, gradient-based methods evaluate the sensitivity of weights via gradients, identifying those whose updates yield negligible effects on the loss function. 
A key advantage of these mathematical heuristics is their general applicability across various neural network architectures, including convolutional neural networks, graph neural networks, and language models.

In the context of language modeling, researchers have sought to characterize the roles of attention heads and the types of linguistic information encoded by language models.
Clark et al.~\cite{clark-etal-2019-bert} found that BERT's attention heads exhibit interpretable patterns aligning closely with syntactic structures, such as direct objects of verbs, determiners, and prepositional objects.
Similarly, Voita et al.~\cite{voita-etal-2019-analyzing} found that attention heads in machine translation models focus on adjacent words, rare tokens, and syntactic dependencies such as  subject-verb relationships.
These findings suggest that individual attention heads can specialize in tracking specific linguistic features.

Motivated by these insights, we propose {\bf Syntactic Attention Pruning (SAP)}, a novel method for effectively pruning attention heads in transformer-based language models by leveraging linguistic information.
The approach is also inspired by human language comprehension, where people can often interpret meaning accurately by focusing on key syntactic cues and grammatical structure, even without perceiving every individual word.
Specifically, SAP leverages {\em syntactic structures} to identify and retain attention heads that are most critical for preserving grammatical and structural integrity.
By prioritizing heads that align strongly with syntactic attributes, SAP ensures essential linguistic functionality is preserved.
Unlike traditional pruning methods based purely on mathematical heuristics, SAP {\em explicitly} connects pruning decisions to linguistic features, thus offering greater interpretability regarding how attention heads contribute to syntactic understanding within the pruned models.

SAP evaluates the importance of each attention head based on its sensitivity to syntactic structures. 
The pruning process involves three main stages.
First, a syntactic analyzer is applied to the target dataset to extract specific syntactic dependencies (e.g., prep, nsubj, aux, etc.; see Section~\ref{sec:dep}) from each sentence.
These dependencies are then aggregated and ranked by their frequency, yielding a hierarchy that reflects the relative importance of syntactic dependencies within the dataset.
In the second stage, the language model is executed on the same dataset.
SAP collects the attention maps from all heads and analyzes their focus across the ranked syntactic dependencies, identifying heads that consistently attend to critical syntactic structures. 
Finally, attention heads that exhibit strong and consistent alignment with top-ranked syntactic dependencies are preserved, while those associated with lower-ranked or infrequent dependencies are pruned.

To further improve robustness, we propose {\bf Candidate Filtering (CF)}, a mechanism that prioritizes heads based on their contribution to model performance.
By evaluating and retaining the most influential heads during inference, CF mitigates performance degradation during pruning.

In summary, this paper makes the following contributions:
\begin{itemize}[itemsep=0pt,nosep]
\item We propose Syntactic Attention Pruning (SAP), a simple yet effective method that leverages syntactic structures to guide attention head pruning. By incorporating linguistic features, SAP improves the interpretability of pruning decisions and the underlying model behavior.
\item We propose Candidate Filtering (CF), a mechanism to improve robustness and reduce the risk of performance degradation during the pruning process.
\item Experimental results show that SAP outperforms existing head pruning methods in a retrain-free setting on the GLUE benchmark.
Analysis of attention maps demonstrates that SAP is more effective than the mathematical metric-based pruning method in preserving heads with richer attention information.
\end{itemize}

The remainder of this paper is organized as follows.
Section~\ref{sec:related} reviews relevant background and related work.
Section~\ref{sec:method} introduces our proposed Syntactic Attention Pruning method and the Candidate Filtering optimization.
Section~\ref{sec:experiment} provides experimental results and analysis. 
Section~\ref{sec:conclusion} concludes the paper.

\section{Background and Related Work} \label{sec:related}
In this section, we first provide background on syntactic dependency.
We then review pruning methods for language models and attention analysis from a linguistic perspective.
Finally, we present the motivation behind our proposed approach.

\subsection{Syntactic Dependency}\label{sec:dep}

\begin{figure}[t!]
    \centering
    \includegraphics[width=.95\linewidth]{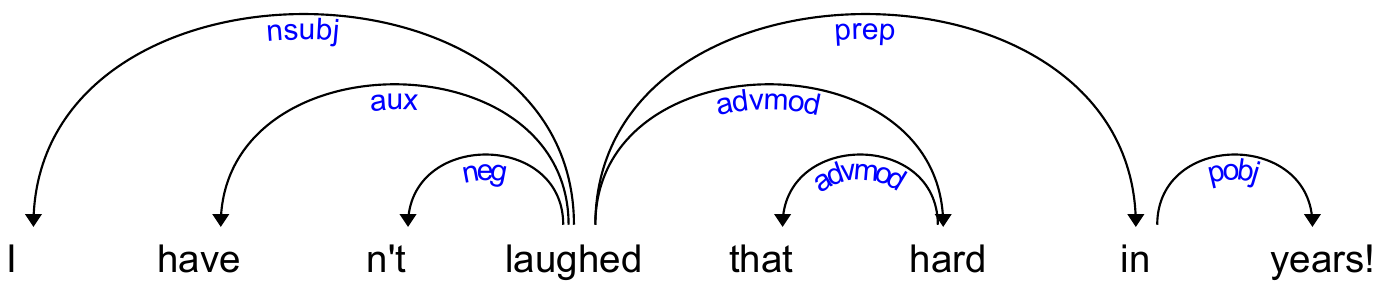}
    \caption{An illustration of syntactic dependencies.}
    \label{fig:exp_dependency}
\end{figure}

Dependency parsing is a fundamental task in natural language processing (NLP) to analyze syntactic relationships between words in a sentence.
It converts a sentence into a structured graph representation $G=(V, A)$, where $V$ is the set of tokens, and $A$ is the set of directed arcs representing syntactic relations between the tokens in $V$.
Each arc links  a {\em head} to a {\em dependent}, with the head serving as the central syntactic unit and the dependent providing supporting information.
Figure~\ref{fig:exp_dependency} shows an example of syntactic dependencies.
The word pair {\tt laughed} $\rightarrow$ {\tt I} is labeled with the dependency {\tt nsubj}, indicating that {\tt I} is the nominal subject of the verb {\tt laughed}.

A widely adopted taxonomy of syntactic relations is developed by the Universal Dependencies (UD) project~\cite{nivre-etal-2016-universal}.
Table~\ref{tbl:dep_label} presents a selection of these dependency labels and their definitions.
For a comprehensive list, please refer to \cite{spacy-dep-list}.
Our experiments utilize the UD labels, and dependency parsing in our experiments is performed using a stack-based shift-reduce parsing algorithm~\cite{10.5555/578789}.
Although the technical details of this method are beyond the scope of this paper, we refer interested readers to the cited reference.

\begin{table}[t!]
    \centering
    \caption{Label and definition of syntactic dependency.}
    \label{tbl:dep_label}
    \resizebox{\linewidth}{!}{
    \begin{tabular}{c|l||c|l}
        \hline
        {\bf Label} & {\bf Definition} & {\bf Label} & {\bf Definition} \\
        \hline
        advmod  & adverbial modifier       & det   & determiner \\
        amod    & adjectival modifier      & dobj  & direct object \\
        appos   & appositional modifier    & hyph  & hyphen \\
        aux     & auxiliary                & mark  & marker \\
        auxpass & auxiliary (passive)      & neg   & negation modifier \\
        case    & case marking             & nmod  & modifier of nominal \\
        cc      & coordinating conjunction & nsubj & nominal subject \\
        ccomp   & clausal complement       & pobj  & object of preposition \\
        conj    & conjunct                 & poss  & possession modifier \\
        csubj   & clausal subject          & prep  & prepositional modifier \\\hline
    \end{tabular}
    }
\end{table}

\subsection{Pruning Methods}

We review two widely used structured pruning methods for transformer-based language models: head pruning and layer pruning.

{\bf Head pruning} removes individual attention heads from the multi-head attention operation in transformer models.
In a pioneering study, Michel et al.~\cite{NEURIPS2019_2c601ad9} introduced mask variables for attention heads.
They used {\em gradients} of these masks as importance scores, estimating the head's sensitivity on model performance.
Their work demonstrated that many of BERT's heads can be removed with minimal impact on downstream task performance.
Alternatively, Voita et al.~\cite{voita-etal-2019-analyzing} proposed using learnable gates (masks) coupled with $L_0$ regularization, encouraging the model to prune less important heads based on training techniques.
Many subsequent studies~\cite{CUI202156,held-yang-2023-shapley,NEURIPS2022_987bed99,liu-etal-2021-ebert} have adopted their framework and developed various loss functions and regularization techniques of different \(L_n\) norms. 

{\bf Layer pruning} removes entire transformer layers, thereby reducing model depth.
Fan et al.~\cite{fan2019reducing} proposed to drop every other layer and showed that this approach works well on small transformers for machine translation, language modeling, and summarization tasks.
Sajjad et al.~\cite{acm-dropping-layers} investigated several pruning strategies, including top-layer, bottom-layer, parameter-based, contribution-based, alternate, and symmetric dropping.
They found that top-layer dropping consistently outperformed other strategies on the GLUE benchmark, achieving strong performance on BERT-based models at 40\% sparsity.

As language models continue to grow in scale, retraining after pruning becomes increasingly expensive, prompting interest in retrain-free pruning techniques.
Several recent studies~\cite{NEURIPS2022_987bed99, wang-etal-2024-pruning, yao-etal-2024-global, dp-prune, park2024accurate} have revisited pruning techniques under this constraint, emphasizing the importance of identifying the most critical heads or layers for removal.
Kwon et al.~\cite{NEURIPS2022_987bed99} proposed a framework that trains learnable masks to guide the removal of attention heads and uses only the downstream dataset to optimize the masks. 
Among these approaches, we compare our SAP against \cite{NEURIPS2022_987bed99}, which represents the current state-of-the-art in retrain-free head pruning.

\subsection{Attention Analysis from a Linguistic Perspective}

With the widespread adoption of the Transformer architecture, researchers have sought to understand the functions of attention heads and types of linguistic information captured by language models.
Clark et al.~\cite{clark-etal-2019-bert} treated each attention head as a basic classifier and evaluated each head's ability to classify various syntactic dependencies.
Their analysis produced three key findings about the BERT model:
1) certain attention heads align with specific syntactic relations, e.g., direct objects of verbs, and determiners of nouns; 
2) attention heads within the same layer exhibit similar behavior and often attend to similar types of syntactic structures.

Voita et al.~\cite{voita-etal-2019-analyzing} examined the functions of attention heads in a pruned model for machine translation.
To evaluate the confidence of individual heads, they introduced a layer-wise relevance propagation (LRP) metric to quantify the confidence with which a head attends to a specific function. 
Their findings indicated that retained heads after pruning tend to correspond to well-defined function categories, such as positional, rare words, and syntactic dependencies.
Morevoer, they observed that as the number of heads is reduced, certain functions shift to other heads, suggesting that these functions are essential and transferable among heads. 

\subsection{Summary}

The aforementioned pruning techniques primarily rely on mathematical metrics to guide pruning decisions.
In contrast, our method leverages syntactic structures as guidance, offering a linguistically grounded approach that more effectively identifies crucial attention heads.
Furthermore, while some previous works have analyzed attention heads from a linguistic perspective, few have attempted to leverage these insights to enhance structured pruning techniques. Inspired by earlier analyses~\cite{clark-etal-2019-bert,voita-etal-2019-analyzing}, we propose SAP, a method that prunes attention heads based on syntactic dependency information. This innovative approach yields substantial improvements over previous works even without retraining, demonstrating the value of linguistic relevance in guiding pruning decisions.

\section{Syntactic Attention Pruning}\label{sec:method}

In this section, we first describe the overall procedure of Syntactic Attention Pruning (SAP).
We then introduce dependency statistics and how they are used to quantify the importance of syntactic dependencies.
Next, we detail the attention ranking process through a running example.
Finally, we present Candidate Filtering, which further enhances the performance of SAP.

\subsection{Overview}

\begin{figure}[t!]
    \centering
    \includegraphics[width=\linewidth]{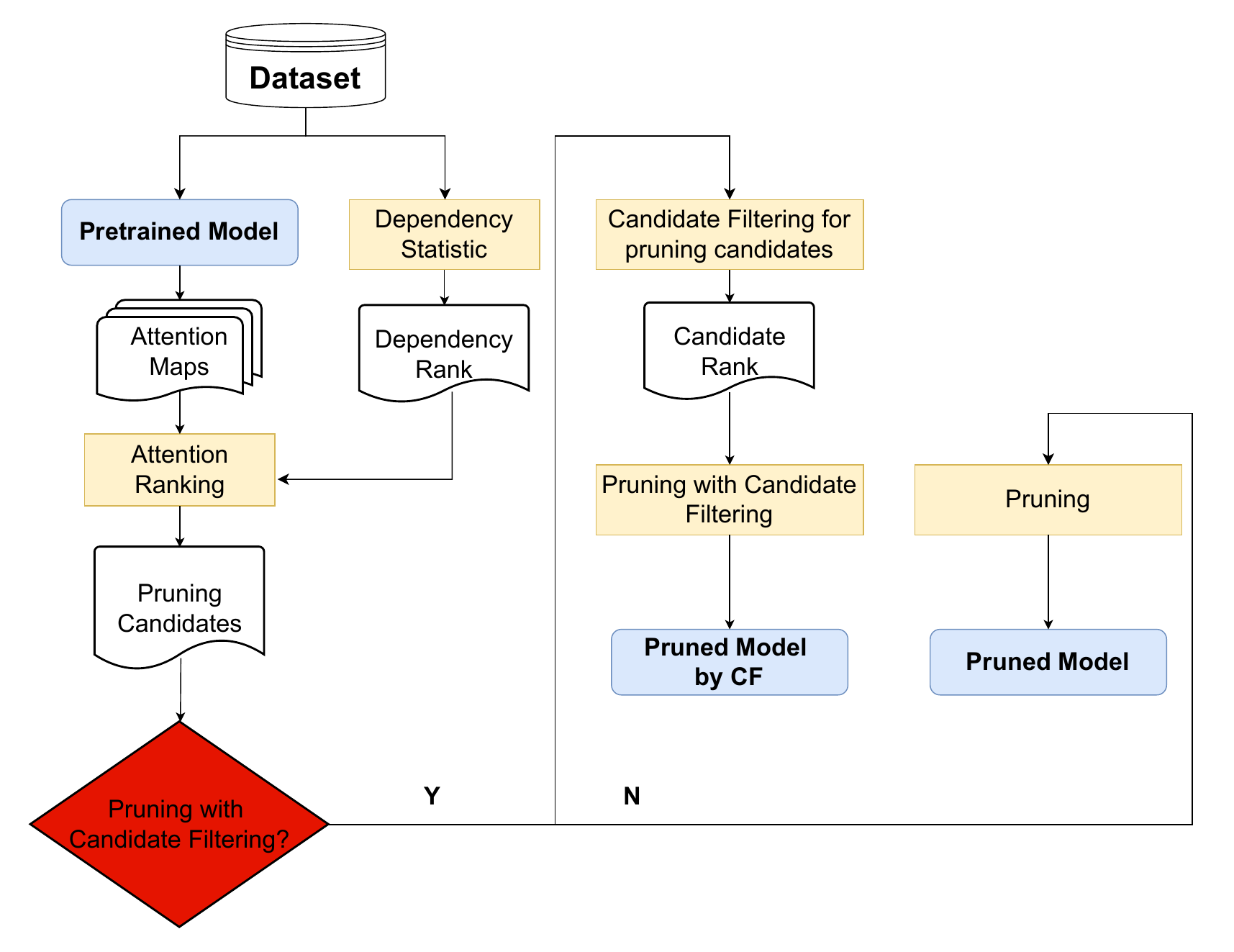}
    \caption{Flow of Syntactic Attention Pruning.}
    \label{fig:SAP_flow}
    \vspace{-5mm}
\end{figure}

We define the structured pruning problem as follows: given a pretrained language model fine-tuned with a downstream dataset, our objective is to identify and prune attention heads that contribute the least to capturing essential syntactic dependencies in the dataset.
Figure \ref{fig:SAP_flow} illustrates the full SAP workflow, with key steps highlighted in yellow.
In the {\em dependency statistics} step (Section~\ref{sec:depstatistic}), we analyze the syntactic dependencies present in the dataset to generate a ranked list of dependency relations based on their frequency.
In the {\em attention ranking} step (Section~\ref{sec:attranking}), we evaluate attention heads using attention maps in conjunction with the top-$k$ syntactic dependencies, and select pruning candidates from the attention ranking list.
To prevent significant performance degradation, we introduce an additional {\em Candidate Filtering (CF)} mechanism (Section~\ref{sec:cf}).
This step ranks the selected candidates by their sensitivities to model performance, ensuring that only the least sensitive heads are pruned.
The resulting pruned model is immediately usable for downstream task inference without retraining. However, light fine-tuning on the downstream dataset can further improve performance. 

\subsection{Dependency Statistic}\label{sec:depstatistic}

\begin{figure*}[t]
    \centering
    \includegraphics[width=\linewidth]{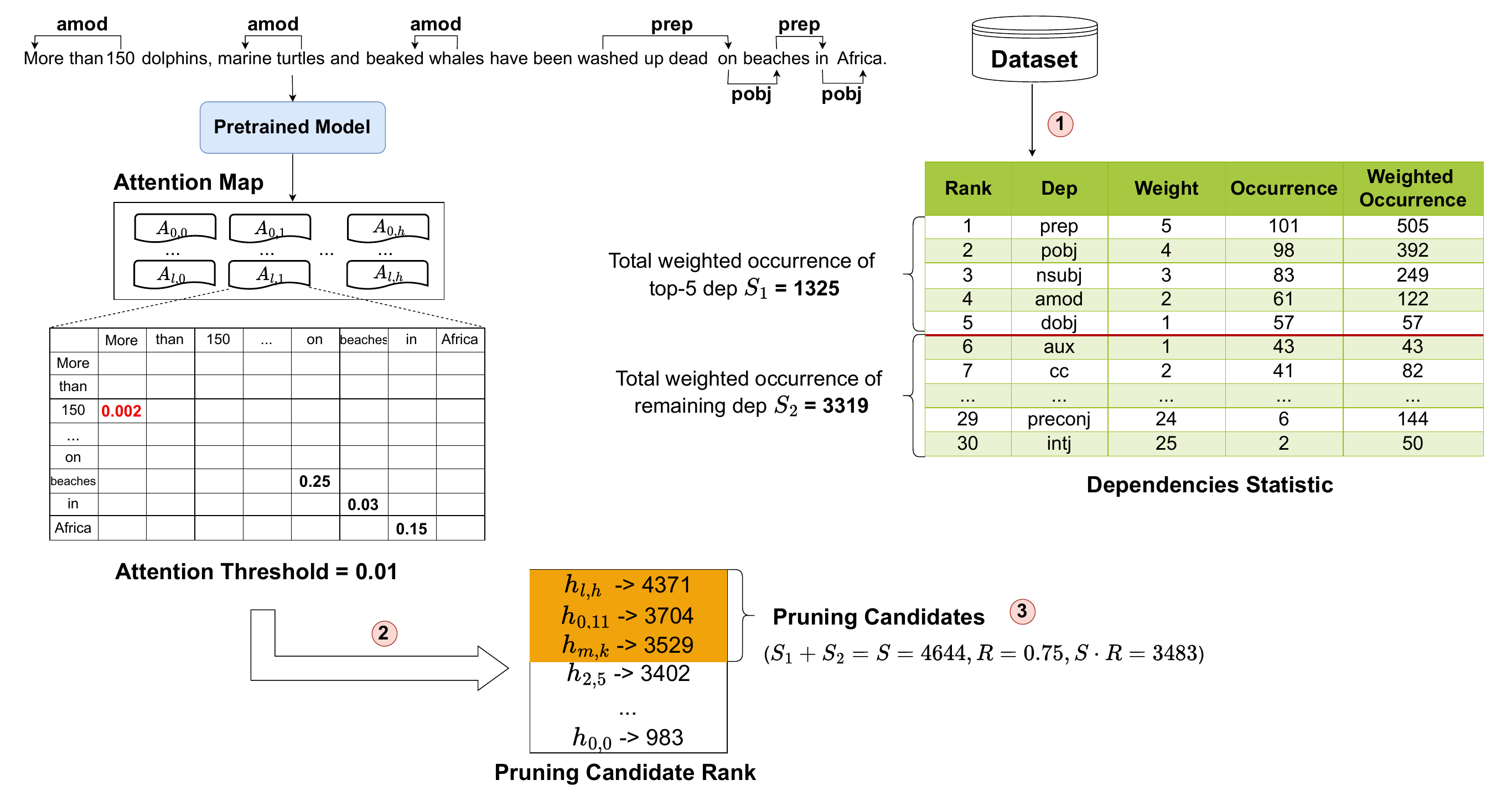}
    \vspace{-6mm}
    \caption{Example of SAP. The sentence is shown with partial syntactic dependencies for brevity.} 
    \label{fig:dep_example}
\end{figure*}

\begin{figure}[t]
    \centering
    \includegraphics[width=\linewidth]{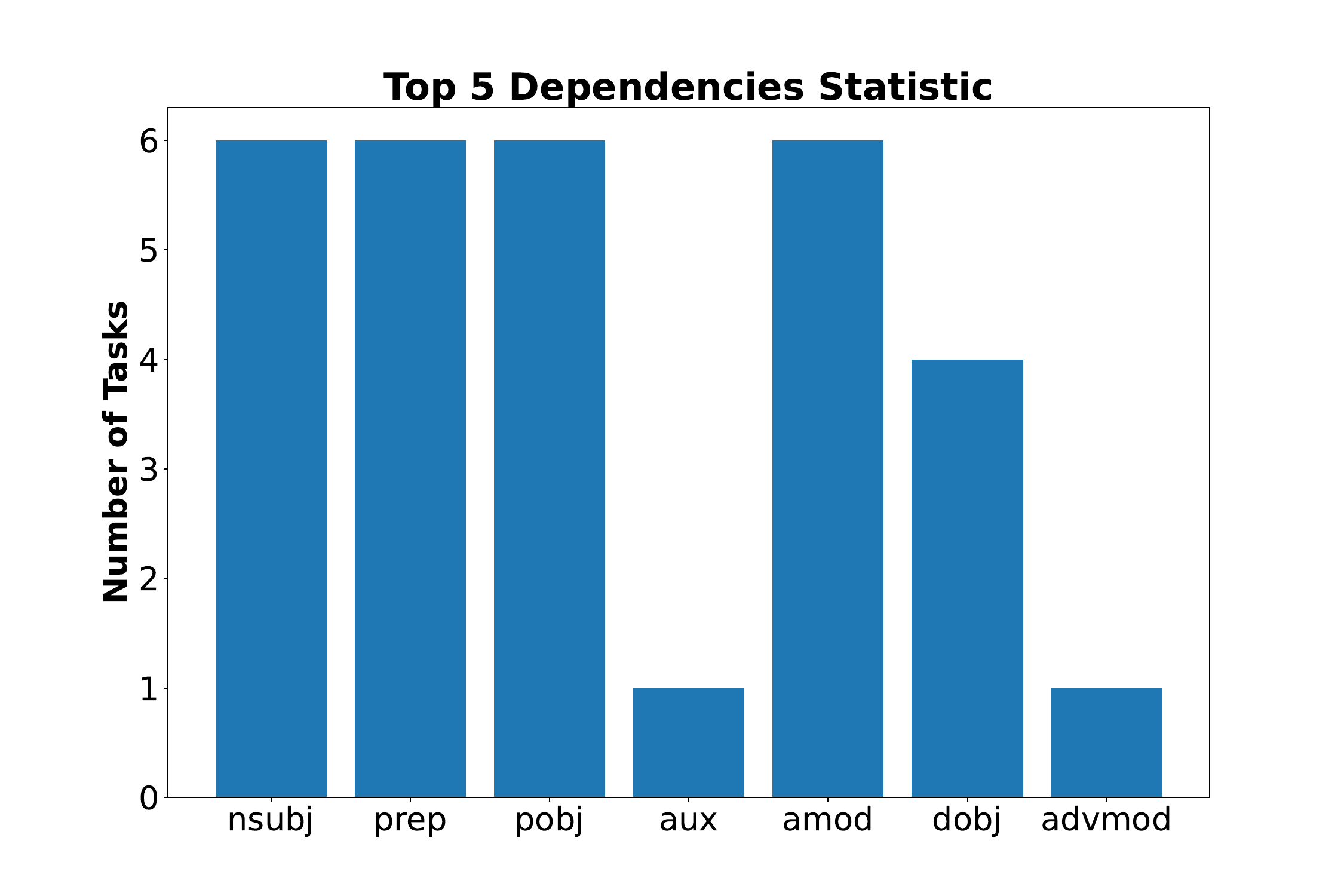}
    \caption{Statistic of top-5 syntactic dependencies in GLUE.
    }
    \label{fig:top_5_dep}
\end{figure}
    
The first step of SAP is to compute dependency statistics for the dataset, aiming to identify the most important syntactic dependencies in the dataset (Step \circled{1} in Figure~\ref{fig:dep_example}).
We begin by performing dependency parsing on each sentence in the dataset, converting it into the syntactic structure described in Section~\ref{sec:dep}.
For example,  in Figure \ref{fig:dep_example}, the sentence ``More than 150 dolphins, marine turtles and beaked whales have been washed up dead on beaches in Africa.'' is taken from the dataset.
After parsing, several syntactic dependencies are extracted: {\tt amod} occurs 3 times, {\tt prep} twice, and {\tt pobj} twice. Other less frequent dependencies are omitted from the figure for clarity. 
We aggregate these counts across the dataset to obtain the total frequency of each syntactic dependency type.
The result of this step is a ranked list of syntactic dependencies, ordered from most to least frequent, which provides guidance for the subsequent SAP stages.
Specifically, given a dataset, a syntactic dependency type is considered {\em important} if its total number of occurrences is ranked within the top-$k$ among all dependency types in the dataset.

To further capture the relative importance of syntactic dependencies, we introduce a rank-based weighting scheme.
That is, the top-$k$ dependency types are assigned weights from $k$ to 1, making higher-ranked dependencies receive greater emphasis.
In contrast, the remaining dependency types (non-top-$k$) are assigned weights in the order from 1 to $(\#dependency\_types-k)$, thus penalizing lower-ranked types more heavily.
By integrating these rank-based weights, SAP can effectively evaluate whether attention heads focus on the most important syntactic dependencies (Section~\ref{sec:attranking}).

Figure~\ref{fig:top_5_dep} shows the top-5 dependency statistics across the GLUE dataset.
The x-axis denotes the dependency types, and the y-axis indicates the number of GLUE tasks in which each dependency type ranks in the task-specific top-5.
We found that the top-5 dependencies for all tasks include the following types: {\tt prep} (prepositional modifier), {\tt nsubj} (nominal subject), {\tt amod} (adjectival modifier), and {\tt pobj} (object of a preposition).
These dependencies commonly involve nouns either functioning as subjects or being modified by other words. 
This observation suggests that crucial task-relevant information in GLUE datasets is often encoded in noun phrases and their modifiers.
Therefore, an effective pruning method should aim to preserve attention heads that attend to these syntactic structures to maintain strong task performance.

\subsection{Attention Ranking}\label{sec:attranking}

After computing the dependency statistics, the next step is to identify attention heads suitable for pruning based on their attention to  syntactic dependencies (Step \circled{2} in Figure~\ref{fig:dep_example}).
The key idea is that an attention head  contributes less to model predictions if it consistently assigns (1) {\em low} attention to the top-$k$ syntactic dependencies, or (2) {\em high} attention to the remaining non-top-$k$ dependencies. 
To assess  this, we reuse the dataset from Step \circled{1}, feed it into the model, and extract attention values from attention maps for word pairs associated with each dependency type.
If a word pair belongs to the group of top-$k$ dependencies and has an attention value {\em below} a certain threshold within a head, we increment the counter for that head, scaled by the weight of the corresponding dependency type.
Similarly, if a word pair belongs to the non-top-$k$ group and receives an attention value {\em above} the threshold, we add the associated weight to the counter for that head.
After processing all sentences of the dataset, we obtain a ranked list of attention heads, where heads with higher counts indicate less attention to key syntactic structures. 
This ranking serves as the basis for selecting potential pruning candidates in the next stage of the SAP pipeline.

The attention threshold is derived directly from the attention maps. Specifically, we aggregate all attention values across the heads and training samples, and take the global average as the threshold.
This threshold is applied uniformly to all attention heads to determine whether a dependency type is considered attended.

Consider the sentence in Figure~\ref{fig:dep_example} as an illustrative example.
The model produces a set of attention maps \({M} \in \mathbb{R}^{L\times H \times n \times n}\), where \(L\) denotes the number of layers, \(H\) the number of attention heads per layer, and \(n\) the number of tokens in the input sentence.
We denote the attention map of head $h$ in layer $l$ as \(A_{l,h} \in \mathbb{R}^{{n}\times{n}}\).
Assume that we are examining the syntactic dependencies and the attention threshold is 0.01.
As the figure shows, the word pair \texttt{More} \textleftarrow \texttt{150} with dependency type \texttt{amond} appears within the top-5 syntactic dependencies.
The attention head assigns a value of 0.002 to this word pair, which falls below the threshold.
Consequently, the corresponding counter \(cnt_{l,1}\) is incremented by {\em two}, according to the weight of the dependency.
In contrast, word pairs with the dependency type {\tt prep} receive attention values above the threshold in this head, so the counter is not updated, suggesting that the head likely captures the {\tt prep} dependency type but not others. Now consider a case where a word pair with dependency type \texttt{preconj} appears in a sentence and receives an attention value above the threshold.
This dependency is ranked 29th, and thus the corresponding counter is incremented by 24.
The same procedure is applied across all attention heads.

In the final step (Step \circled{3}), we select attention heads for pruning based on the ranking derived in  step \circled{2}. 
We define  two parameters:
\begin{itemize}[itemsep=0pt]
\item $S$: the total weighted occurrences of the syntactic dependencies across the entire dataset.
\item $R$: pruning ratio.
\end{itemize}
Given $S$ and $R$, we determine whether an attention head should be pruned using the following criterion:
\begin{equation*}
\textbf{Input}:  X = cnt_{l,h} \in \mathbb{N}_0
\end{equation*}
\begin{equation} \label{eq1}
\textbf{Output}: Y = 
\begin{cases}
            True, & \text{if}\ cnt_{l,h} >=  S\times R \\
            False, & \text{otherwise}
\end{cases}
\end{equation} 
We scan the attention ranking list in descending order of $cnt_{l,h}$, from the highest count to the lowest count.
A head is pruned if its count exceeds the threshold $S \times R$. 
As illustrated in Figure~\ref{fig:dep_example}, the total number of weighted occurrences is $S = 4644$.
Assume $R = 0.75$.
As highlighted in the orange block, the top three heads have counts of 4371, 3704, and 3529, respectively, all exceeding the threshold ($S \times R = 3483$) and thus being selected for pruning.
The parameter \(R\) controls pruning aggressiveness: smaller values allow more heads to be pruned.

\subsection{Pruning with Candidate Filtering}\label{sec:cf}

When users specify a target sparsity ratio, we can adjust $R$ accordingly.
However, if the specified sparsity ratio is too high, pruning too many attention heads may lead to significant performance degradation.
To address this issue, we propose an alternative strategy: instead of providing a sparsity ratio, the user specifies a {\em tolerance threshold}, and SAP automatically determines the maximum number of attention heads that can be pruned without reducing performance below this threshold.
For example, a user may set the tolerance threshold to 90\% of the original dense model's performance.
We call this approach {\em Candidate Filtering (CF)}.

Given the set of head pruning candidates (the orange block in Figure~\ref{fig:dep_example}) from the attention ranking list, CF selects one head at a time, prunes it from the model, and measures the resulting performance degradation relative to the original model. 
After all candidates have been evaluated, they are re-ranked in ascending order of performance degradation, where a smaller degradation indicates the head contributes less to the model performance.
CF then starts over and prunes the original model iteratively.
At each step, it prunes one additional head based on the ranking list, re-evaluates model performance, and stops when performance falls below the specified tolerance threshold.
Additionally, in both SAP and SAP with CF, if the number of pruned heads in a layer exceeds a certain threshold, the entire layer is pruned.

\section{Experiment and Evaluation}\label{sec:experiment}   
\subsection{Experimental Setup}\label{sec:setting}
{\bf Datasets.}
We evaluate our method on the General Language Understanding Evaluation (GLUE) benchmark~\cite{wang2019glue}, a standard dataset in natural language processing.
Specifically, we consider the following tasks: COLA for the corpus of linguistic acceptability,  SST-2 for sentiment analysis with the Stanford sentiment treebank, QQP for Quora Question Pairs, STS-B for the semantic textual similarity benchmark, MNLI for natural language inference,  and QNLI for Question NLI. 

{\bf Models.} 
We evaluate on the pretrained language model BERT$_{BASE}$, which is trained with masked language modeling (MLM).
The model is obtained from HuggingFace and has an architecture of $L$=12 layers, $H$=12 self-attention heads per layer, and hidden size $D$=768.

{\bf Baselines.} 
The SOTA retrain-free head pruning method~\cite{NEURIPS2022_987bed99} is selected as the 
baseline for comparison, which generates masks for both Multi-Head Attention (MHA) and Feed-Forward Network (FFN).
To ensure a fair comparison, we enable only the MHA masks, where attention heads are pruned based on the masks' gradient values until the target sparsity level is reached.

{\bf Implementation of SAP.}
We use spaCy~\cite{honnibal2020spacy} version 3.6.0 to produce dependency statistics for the
dataset.
We use $k$ from 1 to 10 to define the top-$k$ syntactic dependencies as important for all tasks.
Among these, the pruned model corresponding to the best-performing $k$ is selected.
The pruning ratio $R$ is determined based on the user-specified target sparsity.
In our experiments, we use $R=0.5$.
If the number of pruned heads does not reach the target sparsity, we gradually decrease $R$ until the desired sparsity level is achieved.
All experiments are conducted on a machine equipped with an Intel Core i9-12900K CPU and an NVIDIA GeForce RTX 3090 GPU.

\subsection{Performance of Retrain-free Setting}\label{sec:evaluation}
\begin{figure*}[t!]
    \centering
    \label{fig:evaluation_SAP_vs_SOTA}
    \begin{subfigure}[b]{0.3\textwidth}
        \includegraphics[width=\textwidth]{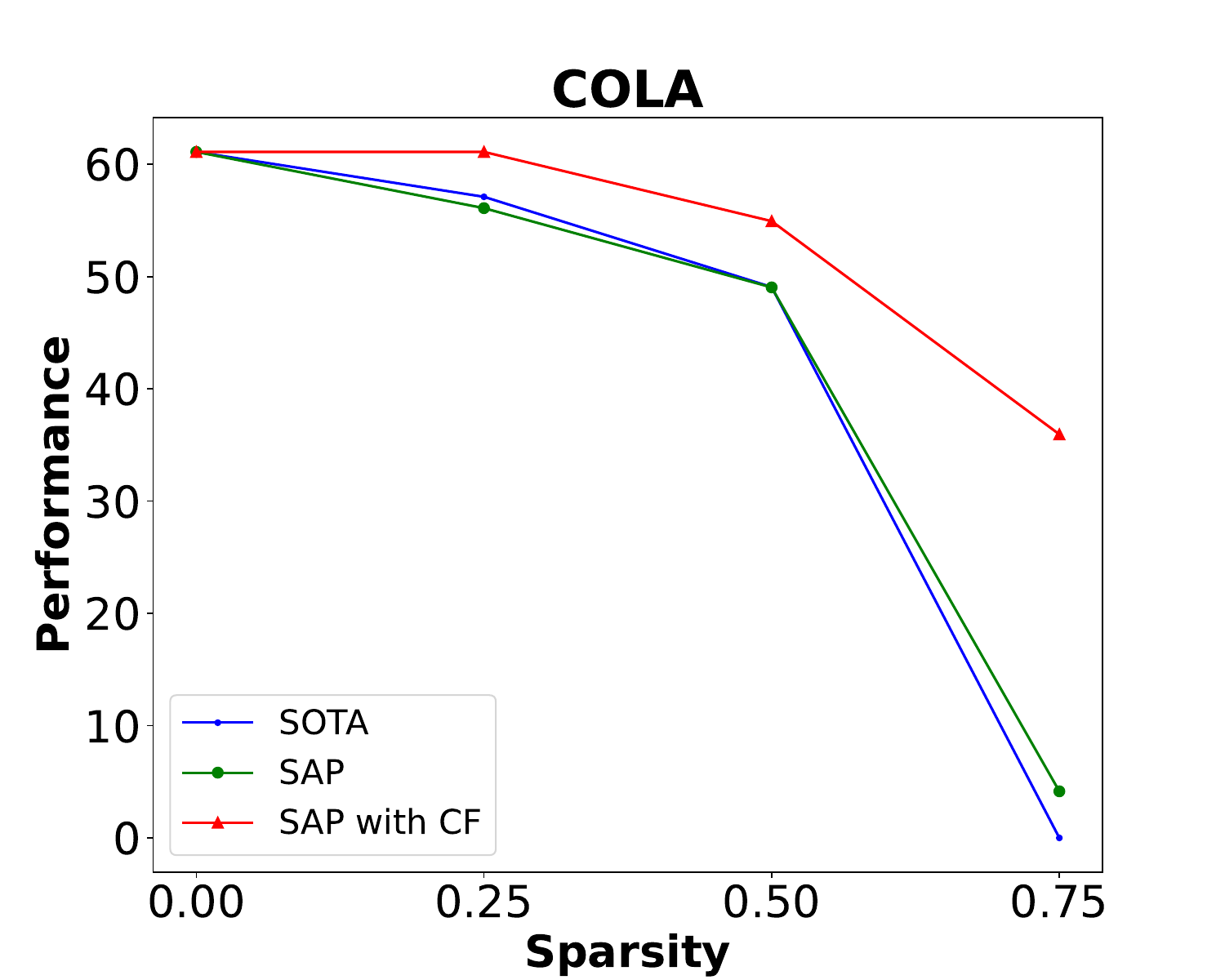}
        \label{fig:R5_cola}
    \end{subfigure}
    \hfill
        \begin{subfigure}[b]{0.3\textwidth}
        \includegraphics[width=\textwidth]{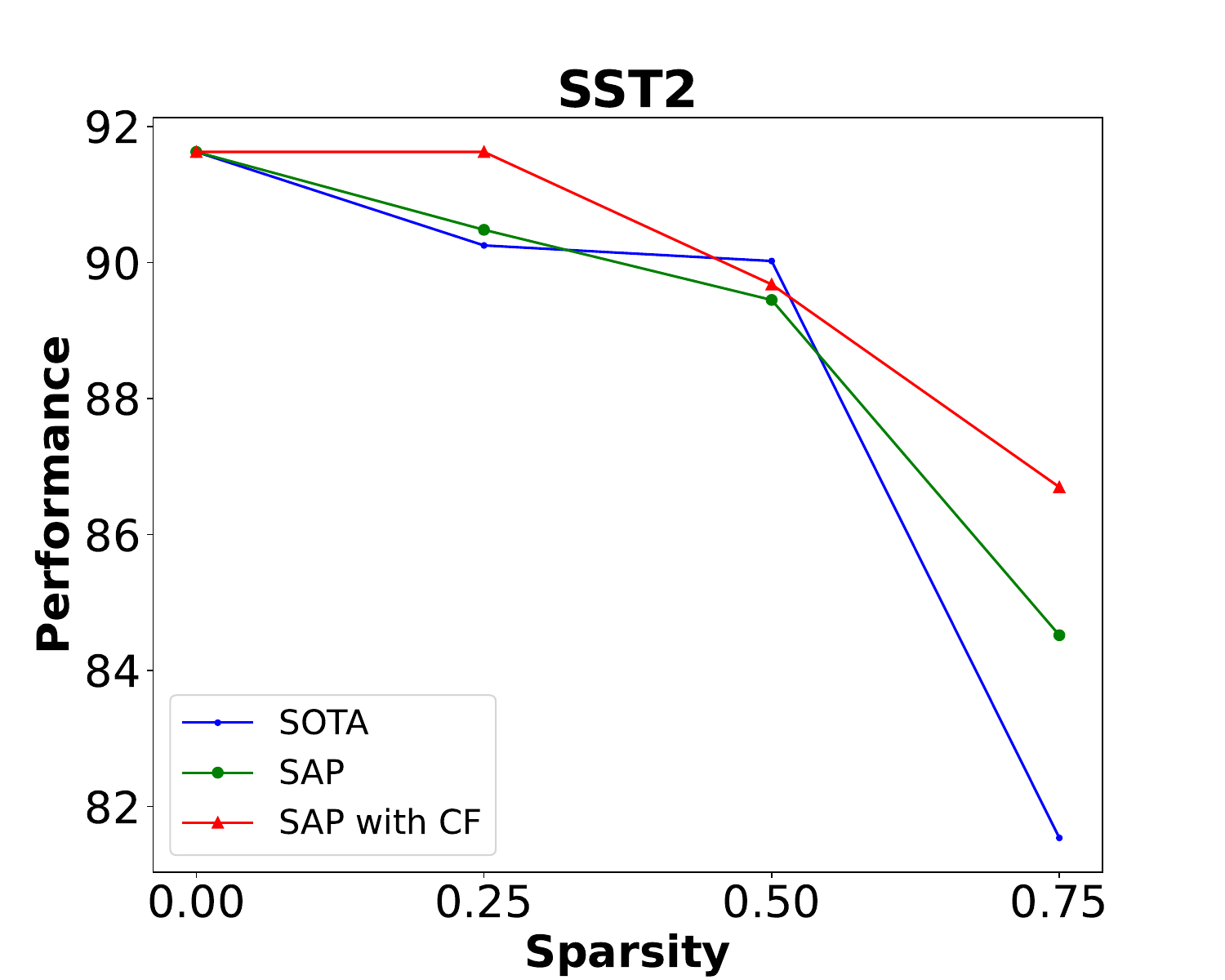}
        \label{fig:R5_sst2}
    \end{subfigure}
    \hfill
    \begin{subfigure}[b]{0.3\textwidth}
        \includegraphics[width=\textwidth]{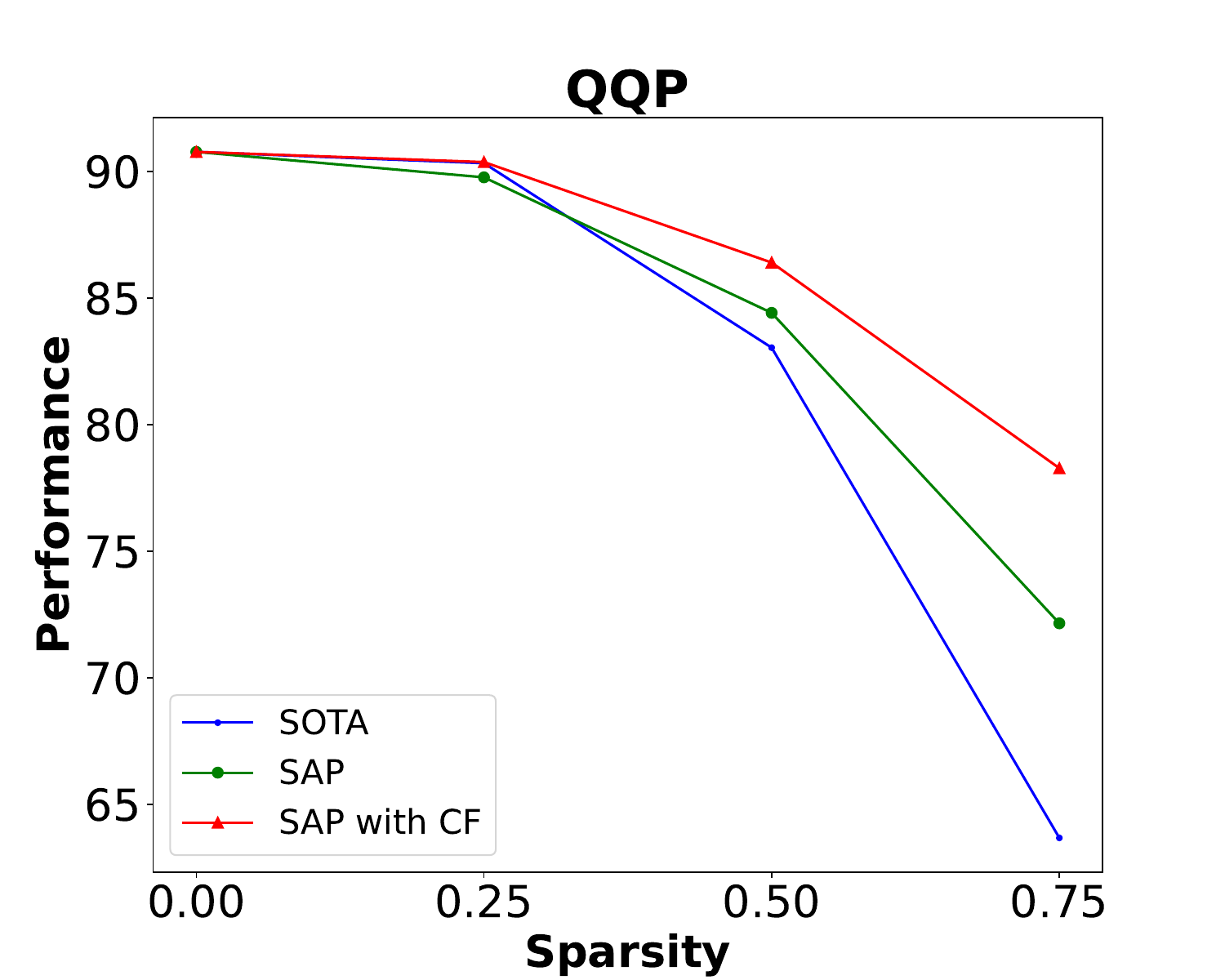}
        \label{fig:R5_qqp}
    \end{subfigure}
    \\
    \begin{subfigure}[b]{0.3\textwidth}
        \includegraphics[width=\textwidth]{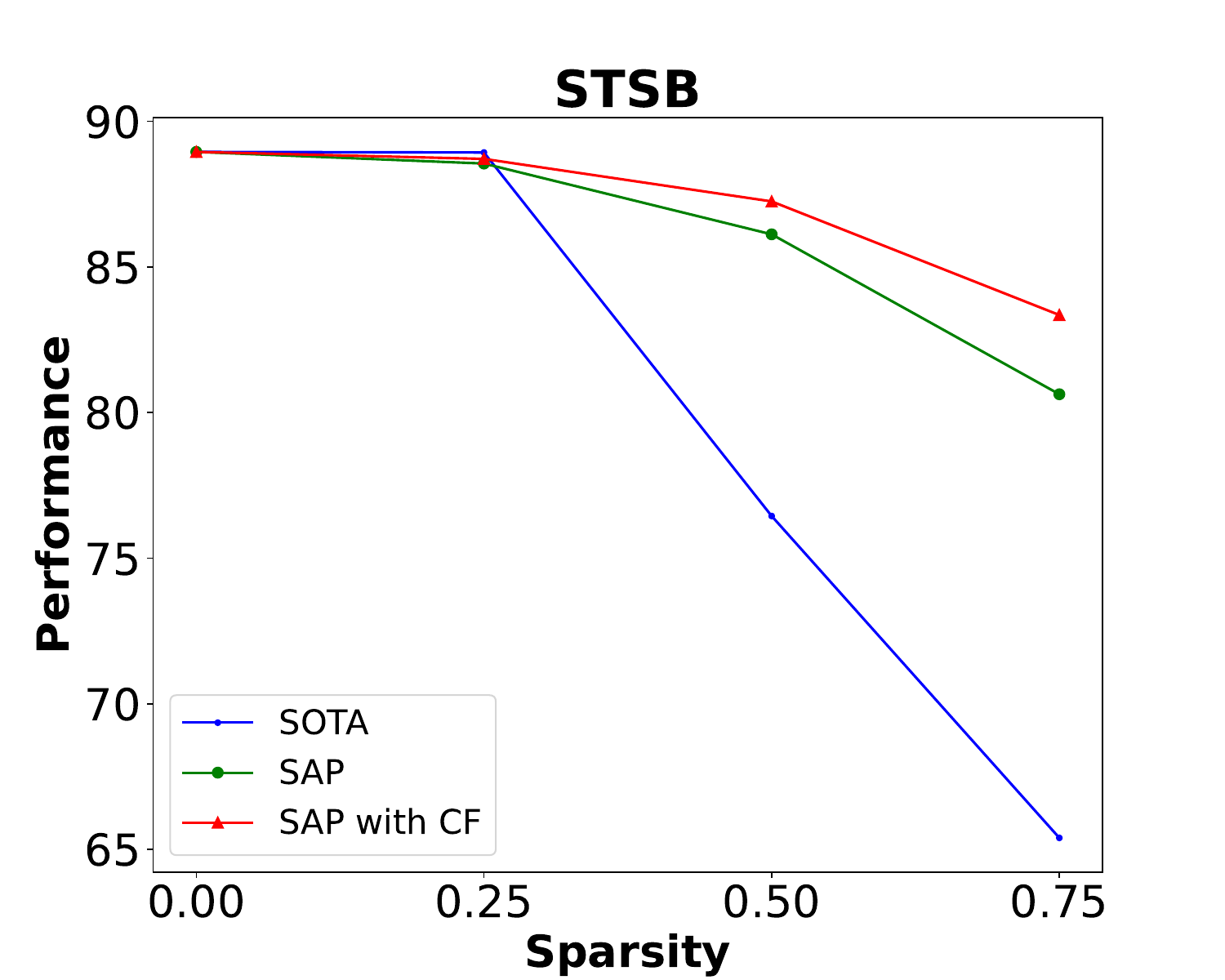}
        \label{fig:R5_stsb}
    \end{subfigure}
        \hfill
    \begin{subfigure}[b]{0.3\textwidth}
        \includegraphics[width=\textwidth]{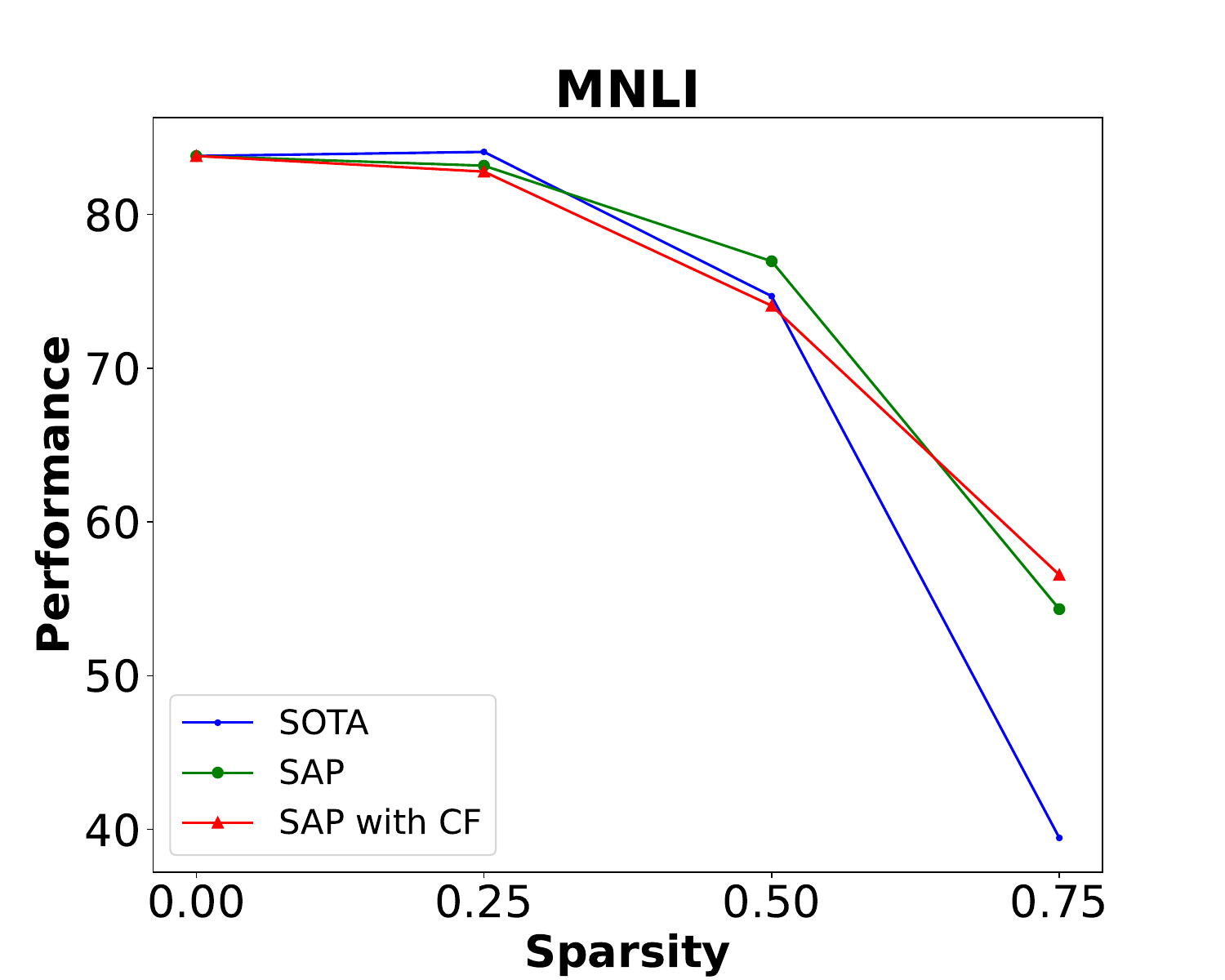}
        \label{fig:R5_mnli}
    \end{subfigure}
    \hfill
    \begin{subfigure}[b]{0.3\textwidth}
        \includegraphics[width=\textwidth]{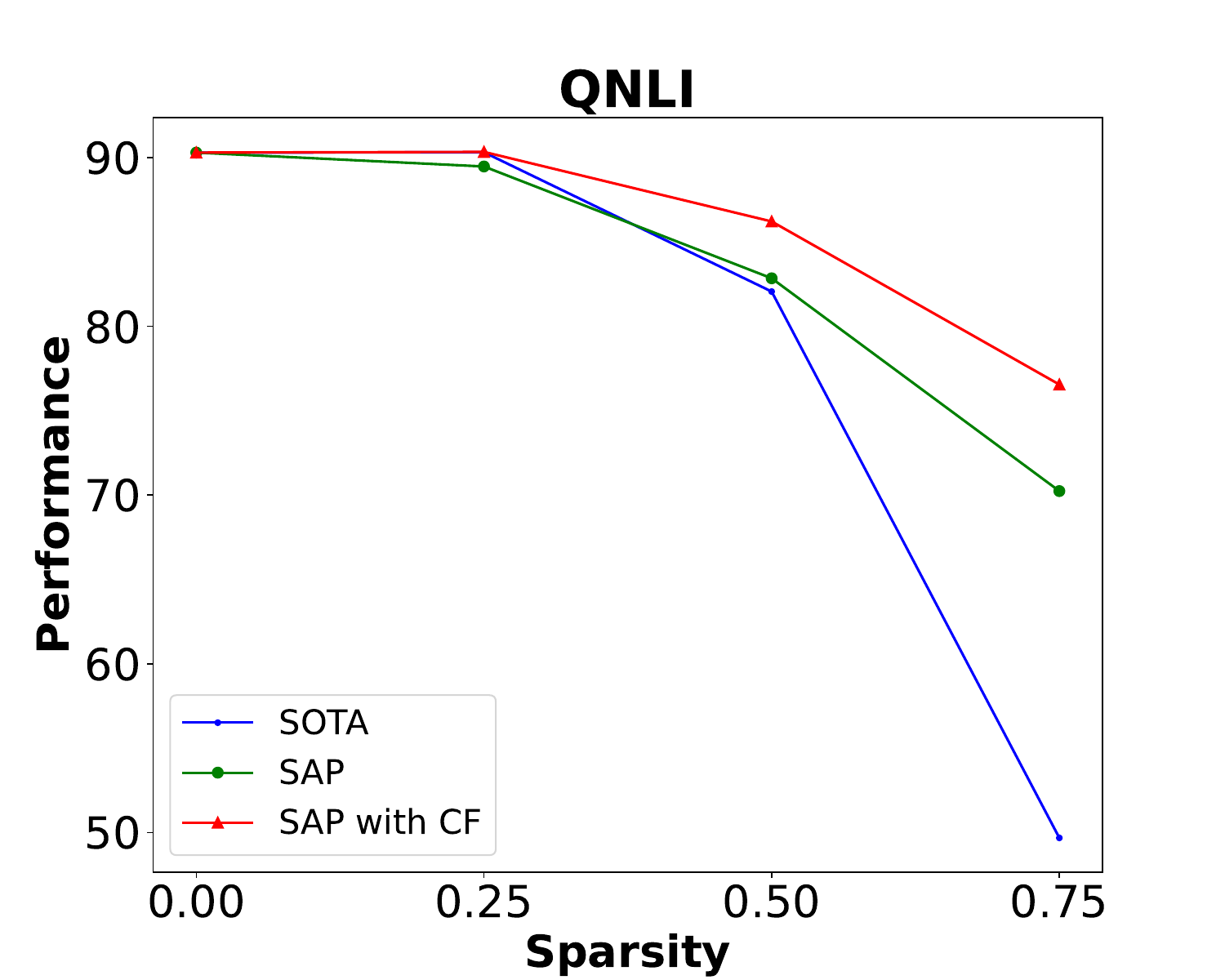}
        \label{fig:R5_qnli}
    \end{subfigure}
    \vspace{-2mm}
    \caption{Performance comparison at different sparsity levels.}
\label{perf_reduction}
\end{figure*}

\begin{figure*}[]
    \centering
    \includegraphics[width=0.8\textwidth]{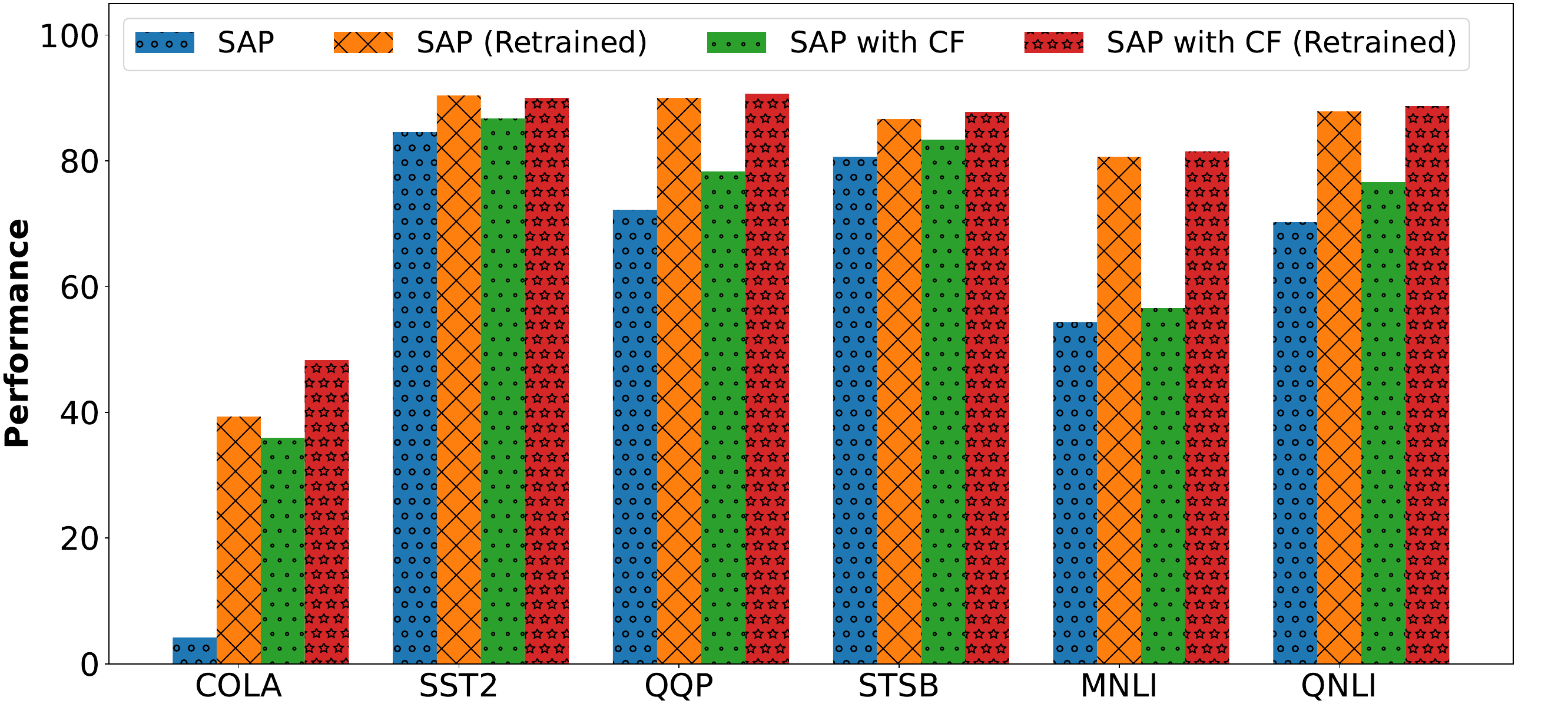}
    \caption{Performance enhancement after retraining.}
    \label{fig:perf_retrain_free_vs_retrained}
\end{figure*}

Figure~\ref{perf_reduction} compares the performance of our method with the baseline under different sparsity levels.
The results highlight several key observations.
First, the performance of models pruned by SAP exhibits flatter curves than those pruned by the baseline method, suggesting that pruning guided by syntactic dependencies is more effective to identify important heads than relying solely on mathematical criteria.
Second, as sparsity increases, the performance gap between SAP and the baseline becomes more pronounced, demonstrating that SAP is more resilient to performance degradation.
Third, combining SAP with CF further improves performance compared to standalone SAP.
SAP provides coarse-grained pruning by filtering out heads with low syntactic relevance, while CF refines the process through fine-grained self-performance degradation ranking.
Overall, the proposed SAP outperforms the SOTA approach while offering stronger resilience, especially at higher sparsity levels.

While SAP demonstrates strong performance in the retrain-free setting, the pruned models can be further enhanced with a few additional epochs of retraining.
Figure~\ref{fig:perf_retrain_free_vs_retrained} illustrates the impact of retraining at 0.75 sparsity.
The results show that both SAP and SAP+CF exhibit substantial gains, with consistent improvements across all tasks.

\subsection{Visualization of Attention Maps}

\begin{figure*}[t]
    \vspace{0.5mm}
    \centering
    \begin{subfigure}[b]{\textwidth}
        \includegraphics[width=\textwidth]{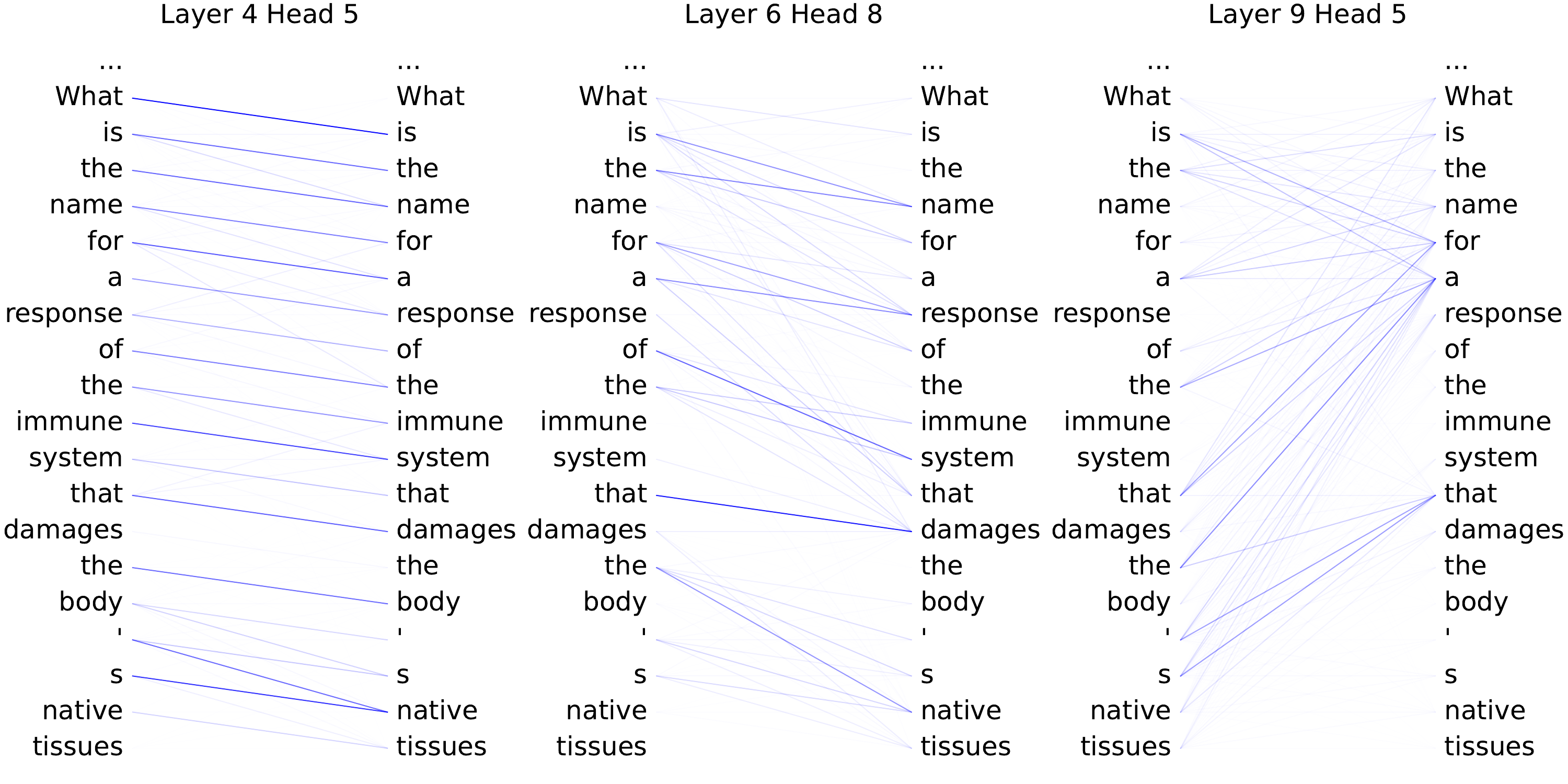}
        \caption{SOTA}
        \label{fig:attnMap_SOTA}
    \end{subfigure}
    \hfill
    \begin{subfigure}[b]{\textwidth}
        \includegraphics[width=\textwidth]{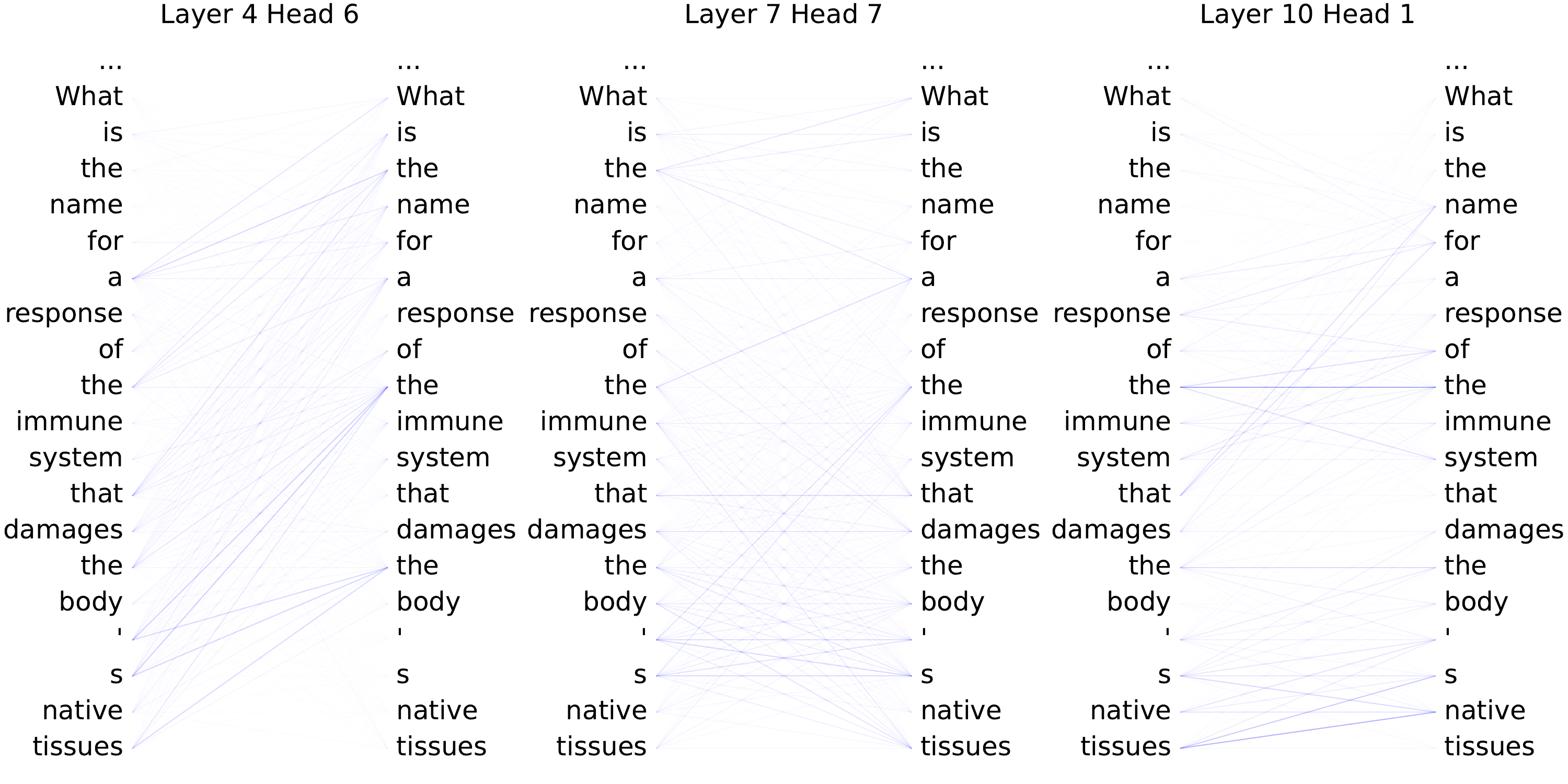}
        \caption{SAP}
        \label{fig:attnMap_SAP}
    \end{subfigure}
    \caption{Comparison of Attention Maps between SOTA vs. SAP on a QNLI Example}
\end{figure*}

This section analyzes the attention maps of heads pruned by SAP and the baseline. We examine the heads pruned by both methods on the QNLI task.
Figures \ref{fig:attnMap_SOTA} and \ref{fig:attnMap_SAP} show the attention maps of a subset of pruned heads by SOTA and SAP, respectively.
In each map, a blue line connects a word pair representing a syntactic dependency, where the word on the left is the dependent and on the right is the head.
Line intensity reflects the corresponding attention value, with darker lines indicating higher attention.
Comparing the attention maps, we observe that heads pruned by SOTA generally exhibit more dark lines than those pruned by SAP, meaning that more important heads and syntactic dependencies are pruned than SAP.
This demonstrates that our method, which prunes based on attention values of syntactic dependencies, is more effective at preserving heads with richer attention information than the mathematical metric–based pruning method.

\subsection{Effect of $k$}

\begin{figure}[t]
    \vspace{0.5mm}
    \centering
    \label{fig:hyperparas}
    \begin{subfigure}[b]{0.45\textwidth}
        \includegraphics[width=\textwidth]{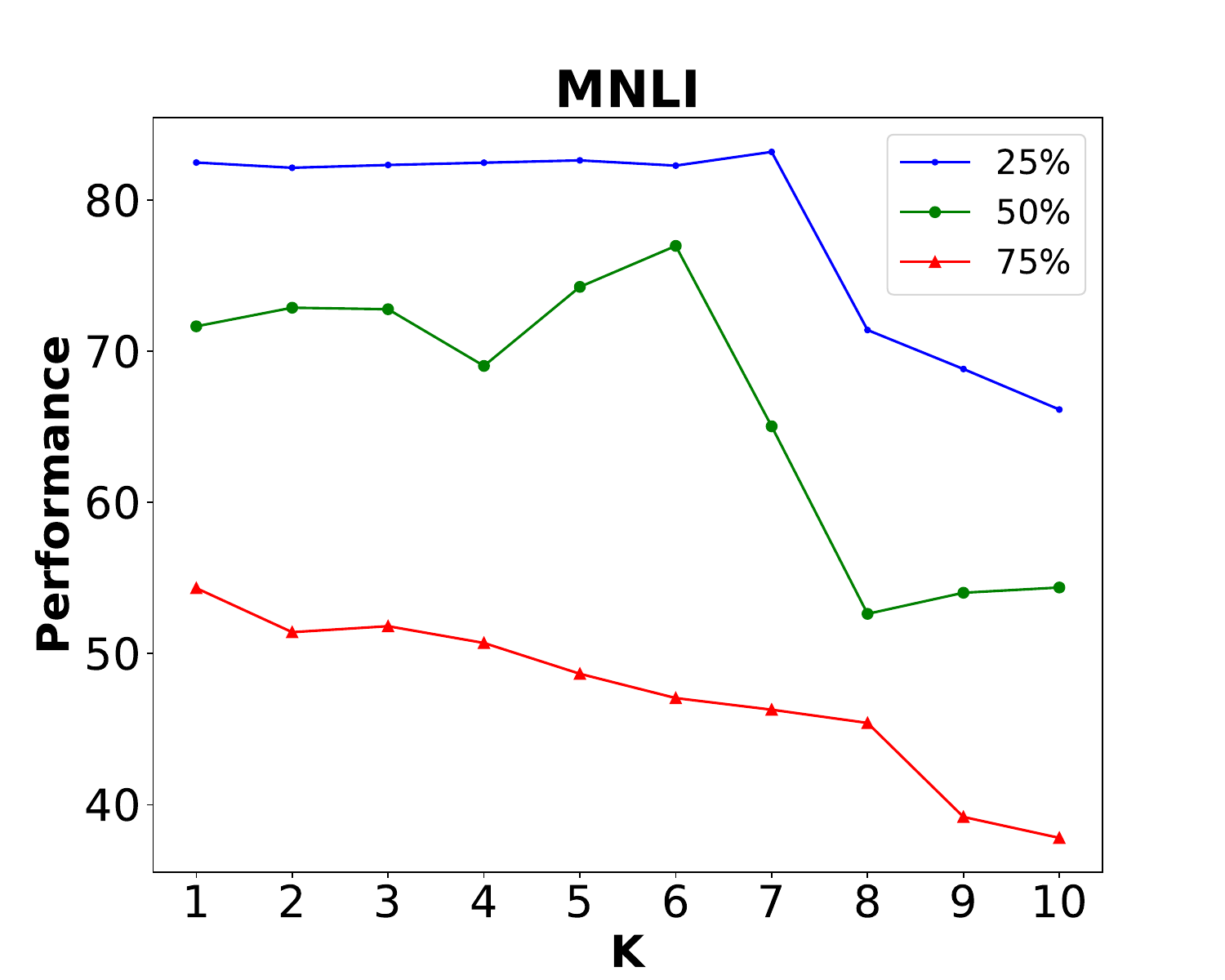}
        \caption{MNLI}
        \label{fig:topk_mnli}
    \end{subfigure}\\
    \begin{subfigure}[b]{0.45\textwidth}
        \includegraphics[width=\textwidth]{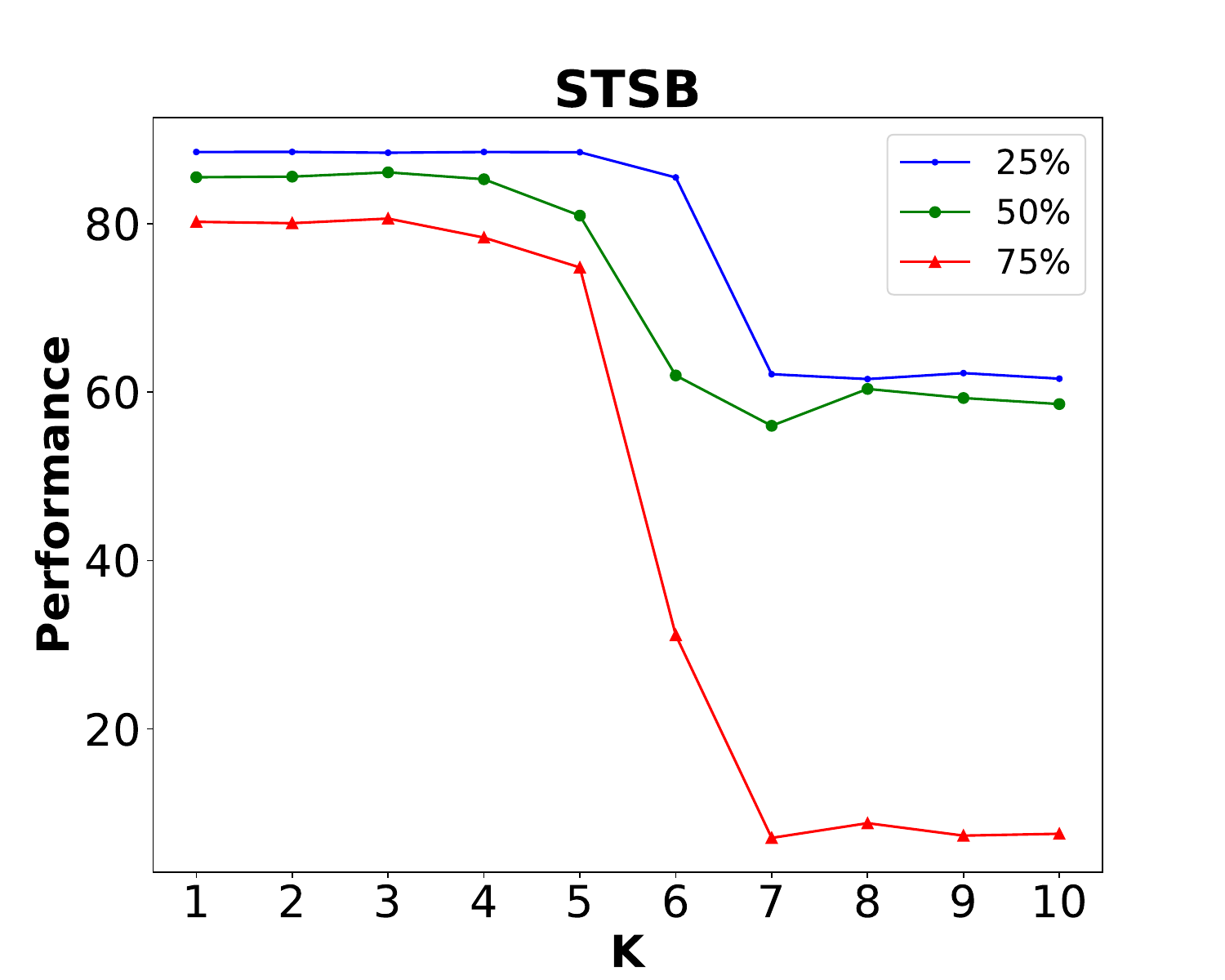}
        \caption{STS-B}
        \label{fig:topk_stsb}
    \end{subfigure}
    \caption{Performance of top-X syntactic dependencies.}
\end{figure}

This section examines the effect of $k$ on the selection of important syntactic dependencies.
We vary the $k$ values from 1 to 10 and conduct experiments on two GLUE tasks.
Figures \ref{fig:topk_mnli} and \ref{fig:topk_stsb} show the performance evaluated at sparsity levels 0.25, 0.5, and 0.75 in the retrain-free setting for the MNLI and STS-B benchmarks, respectively.
The results indicate that the best performance is generally achieved when $k \in [3, 7]$, whereas performance degrades as $k$ becomes either smaller or larger.
Similar trends are observed across other GLUE benchmark tasks.
This can be explained by the effect of $k$ on the pruning criteria: smaller $k$ values make the criteria too loose, preventing the system from distinguishing the relative importance among heads, whereas larger $k$ values make the pruning criteria too strict, causing more heads that focus on relatively important syntactic dependencies to be removed.
Consequently, choosing a $k$ value outside this range in SAP is likely to degrade model performance.

\section{Conclusion}  \label{sec:conclusion}    

This paper introduces Syntactic Attention Pruning (SAP), a novel head pruning method for Transformer-based language models that integrates linguistic insights into the pruning process.
SAP leverages syntactic dependency structures to guide the identification and removal of redundant attention heads.
To improve pruning robustness, we further introduce Candidate Filtering (CF), a mechanism that prioritizes heads according to their contribution to model performance and mitigates quality degradation during pruning.
Our empirical evaluation shows that SAP effectively preserves critical heads with strong syntactic relevance and outperforms the state-of-the-art mathematical metric-based pruning methods in retrain-free settings.
Given its effectiveness, SAP provides a promising foundation for future research in model compression and can be applied to other Transformer-based language models.
Nevertheless, SAP has a limitation: its effectiveness relies on distinguishing between the syntactic dependencies.
In scenarios where many heads attend to diverse syntactic relations, the method may be less effective at distinguishing their relative importance.
We plan to address this challenge in the future for further enhancing the efficiency and applicability of SAP.

\printbibliography

\appendix

\end{document}